\documentclass[runningheads]{llncs}

\usepackage{eccv}

\usepackage{eccvabbrv}

\usepackage{graphicx}
\usepackage{booktabs}

\usepackage[accsupp]{axessibility}  %

\usepackage{hyperref}

\usepackage{orcidlink}

\usepackage{algorithm}
\usepackage{algpseudocode}
\usepackage{booktabs}
\usepackage{bm}
\usepackage{caption}
\usepackage{colortbl}
\usepackage{dsfont}
\usepackage{enumitem}
\usepackage{float}
\usepackage{graphicx}
\usepackage{hyperref}
\usepackage{lipsum}
\usepackage{listings}
\usepackage{mathtools} 
\usepackage{multicol}
\usepackage{multirow}
\usepackage{pifont}
\usepackage{wrapfig}
\usepackage{subcaption}
\usepackage[normalem]{ulem}
\usepackage{url}
\usepackage{xcolor}
\newcommand{\cmark}{\ding{51}}

\def\1{\bm{1}}

\usepackage{xcolor}
\usepackage{soul}
\usepackage{amsmath}
\usepackage{mathabx}

\usepackage{amsmath}

\begin{document}

\title{\textsc{RePL}: Pseudo-label Refinement for Semi-supervised LiDAR Semantic Segmentation}

\titlerunning{Pseudo-label Refinement for Semi-supervised LiDAR Semantic Segmentation}

\author{Donghyeon~Kwon\inst{1}\textsuperscript{*}\orcidlink{0009-0006-2791-1471}
    \and Taegyu~Park\inst{2}\textsuperscript{*}\orcidlink{0009-0006-6387-644X}
    \and Suha~Kwak\inst{2,3}\orcidlink{0000-0002-4567-9091}}

\authorrunning{D.~Kwon et al.}

\institute{\textsuperscript{1}GENGENAI, Republic of Korea
    \quad \textsuperscript{2}Dept. of CSE, POSTECH, Republic of Korea
    \\ \textsuperscript{3}Graduate School of AI, POSTECH, Republic of Korea
    \\ \email{\inst{1}kinux98@gmail.com,
        \quad \inst{2}\{taegyu.park,suha.kwak\}@postech.ac.kr}}

\maketitle
{\renewcommand{\thefootnote}{*}
    \begin{NoHyper}\footnotetext{Equal contribution.}\end{NoHyper}
}\vspace{-4mm}

\begin{abstract}
    Semi-supervised learning for LiDAR semantic segmentation often suffers from error propagation and confirmation bias caused by noisy pseudo-labels.
    To tackle this chronic issue, we introduce \textsc{RePL}, a novel framework that enhances pseudo-label quality by identifying and correcting potential errors in pseudo-labels through masked reconstruction, along with a dedicated training strategy.
    We also provide a theoretical analysis demonstrating the condition under which the pseudo-label refinement is beneficial, and empirically confirm that the condition is mild and clearly met by \textsc{RePL}.
    Extensive evaluations on the nuScenes-lidarseg and SemanticKITTI datasets show that \textsc{RePL} improves pseudo-label quality substantially, and in consequence, achieves the state of the art in semi-supervised LiDAR semantic segmentation.
    \vspace{-1ex}
\end{abstract}

\section{Introduction}
\vspace{-1ex}
Outdoor LiDAR semantic segmentation, the task of assigning semantic labels to every point in outdoor 3D scenes, plays a crucial role in diverse applications such as autonomous driving~\cite{autodriving0, autodriving1, autodriving3} and robotics~\cite{robot2, robot1}. Recent progress in this field has been largely driven by supervised learning on large-scale point cloud datasets~\cite{semantickitti, nuscenes-lidarseg}. However, collecting dense annotations for 3D point clouds is prohibitively costly and
time-consuming,
which limits the scale and class diversity of training data. To alleviate this bottleneck, a large body of research has explored data-efficient training paradigms such as semi-supervised learning~\cite{jiang2021guided, kong2023lasermix, li2023less, li2024density, liu2024ittakestwo, liu2025exploring}, weakly supervised learning~\cite{weak3,weak4}, and unsupervised learning~\cite{unsup3,unsup4}. In this work, we study semi-supervised learning for LiDAR semantic segmentation, where only a subset of 3D scenes for training is manually labeled and the remainder is unlabeled.

At the core of semi-supervised learning lies the challenge of leveraging unlabeled data effectively for training. To this end, existing methods for LiDAR semantic segmentation commonly leverage consistency regularization~\cite{kong2023lasermix, liu2024ittakestwo, liu2025exploring} and contrastive learning~\cite{jiang2021guided, li2024density, liu2024ittakestwo}. Consistency regularization encourages stable and invariant predictions by enforcing similar outputs under different perturbations of the same input, while contrastive learning promotes feature representations that bring samples of the same pseudo-labels closer and push those of different pseudo-labels farther apart.
Although these approaches often yield substantial gains, they have a fundamental vulnerability, confirmation bias towards erroneous pseudo-labels~\cite{confirmation5,  confirmation2}, since they blindly exploit the model's predictions as pseudo-labels for training the model with unlabeled data; this could cause performance to deteriorate as training progresses.

\begin{figure}[t]
    \centering
    \includegraphics[width=\linewidth]{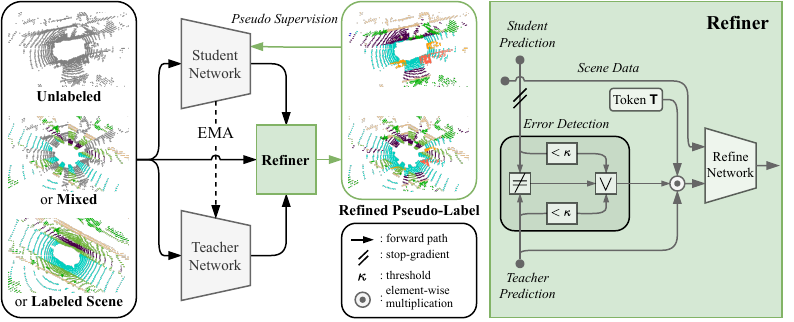}
    \caption{
        Overview of \textsc{RePL}.
        The teacher generates predictions for unlabeled LiDAR scenes, which are used as pseudo-labels for the student, and is updated via exponential moving average (EMA) of the student. The pseudo-label refiner detects erroneous pseudo-labels by confidence-based agreement between the teacher and student, and then corrects them through masked reconstruction with learnable tokens. The final refined pseudo-labels combine reliable teacher predictions with reconstructed ones, yielding improved supervision for semi-supervised learning of the student.}
    \label{fig:motivation}
    \vspace{-3ex}
\end{figure}

To handle noisy pseudo-labels, recent studies
have proposed confidence-based filtering~\cite{kong2023lasermix,li2023less,liu2025exploring}, which discards predictions with low confidence under the assumption that such labels are more error-prone. Loss reweighting methods~\cite{li2024density,liu2024ittakestwo} take a complementary approach by retaining all pseudo-labeled samples but adjusting their contribution through confidence weighted loss scaling. While both strategies assuage the impact of unreliable labels, they remain post-hoc in nature as they adjust sample utilization only after pseudo-labels have been assigned, rather than improving their quality at the point of generation.

To tackle this issue, we introduce a novel pseudo-label refinement framework, named \textsc{RePL}~(\textbf{Re}finement of \textbf{P}seudo-\textbf{L}abels), which is illustrated in \cref{fig:motivation}.
\textsc{RePL} integrates two key components: teacher-student networks for LiDAR semantic segmentation~\cite{MeanTeacher}
and a pseudo-label refiner that identifies errors on pseudo-labels and corrects them. The teacher network offers initial predictions for LiDAR scenes, which are in turn used as pseudo-labels for training the student with unlabeled data; the teacher is then updated by exponential moving average (EMA) of the student.
The pseudo-label refiner identifies potentially unreliable regions in the pseudo-labels and reconstructs them into cleaner supervisory signals for the student.
Specifically, the refiner operates through
two stages.
First, it estimates potentially erroneous pseudo-labels using a simple confidence-based agreement between teacher and student predictions. Second, it reconstructs these pseudo-labels through a process inspired by masked autoencoders~\cite{mae}; in this step, unreliable areas are replaced with mask tokens, and the refiner reconstructs them to produce refined predictions. Even with a simple error estimation strategy, \textsc{RePL} achieves significant performance improvements, while remaining extensible to more sophisticated error detection methods.

The success of our method heavily depends on the ability of the pseudo-label refiner.
However, training the refiner in the semi-supervised learning setting is challenging
due to the scarce supervision: the supervision derived from the discrepancies between predicted and ground-truth labels is available only for a small subset of training data.
We address this issue in two ways.
First, during training the refiner, we apply additional random masking to make its reconstruction more challenging. This forces the refiner to develop a better contextual understanding rather than simply memorizing patterns.
Second, we mix 3D scenes, augmenting labeled scenes with unlabeled ones before feeding them to the segmentation networks. This produces strongly augmented segmentation predictions with higher and diverse error rates, and enables the refiner to partially experience prediction errors for unlabeled scenes.
We also provide a theoretical analysis demonstrating the condition under which the refinement is beneficial, and empirically confirm that the condition is mild and clearly met by \textsc{RePL}.

Following the established practice~\cite{kong2023lasermix,liu2024ittakestwo, liu2025exploring}, we evaluated our method on the nuScenes-lidarseg~\cite{nuscenes-lidarseg} and SemanticKITTI~\cite{semantickitti} benchmarks while varying the ratio of labeled data, where it demonstrated significant performance improvements over the supervised learning baseline
and outperformed latest methods for semi-supervised learning. In summary, our contribution is three-fold:

\begin{itemize}
    \item We propose a semi-supervised LiDAR semantic segmentation framework, dubbed \textsc{RePL}, that refines pseudo-labels via error estimation and masked reconstruction.
    \item We provide a theoretical analysis establishing the condition under which pseudo-label refinement improves upon the teacher-only baseline.
    \item \textsc{RePL}
          achieved the state of the art on the two public benchmarks.
\end{itemize}
\label{intro}
\section{Related Work}

\paragraph{LiDAR Semantic Segmentation.}
LiDAR semantic segmentation assigns semantic labels to each point in large-scale point clouds~\cite{cylindernet,rangenet++}. Early work transformed raw 3D scans into 2D representations such as range images or bird's-eye-view maps to reuse established 2D convolutional backbones. RangeNet++~\cite{rangenet++} and SalsaNext~\cite{salsanext} demonstrated that simple projection can yield competitive results. Another research stream discretized 3D space into regular or cylindrical grids, as seen in PolarNet~\cite{polarnet}, Cylinder3D~\cite{cylindernet}, and sparse convolutional frameworks like MinkowskiNet~\cite{minkowskinet}. Meanwhile, methods directly processing raw points gained traction, from PointNet~\cite{pointnet} to RandLA-Net~\cite{randla} and stratified architectures~\cite{stratified2, stratified}, capturing both local details and long-range context. Despite performance gains, these methods require dense manual annotations, which are costly and not scalable.

\paragraph{Semi-supervised LiDAR Semantic Segmentation.}
To address this annotation bottleneck, semi-supervised learning leverages limited labeled data along with abundant unlabeled point clouds~\cite{kong2023lasermix,liu2024ittakestwo}.
Earlier work improved pseudo-label reliability: GPC~\cite{jiang2021guided} used confidence thresholds to reduce error propagation, LaserMix~\cite{kong2023lasermix} exploited spatial priors of LiDAR beams, enforcing prediction consistency through beam mixing across scans, and Lim3D~\cite{li2023less} employed a memory bank for contrastive learning to alleviate class imbalance.
Recent methods introduced richer constraints: DDSemi~\cite{li2024density} employed density-guided contrastive learning with dual-space hardness sampling for sparse regions, AIScene~\cite{liu2025exploring} addressed intra-scene inconsistency through patch-based mixing at scene and instance levels, and IT2~\cite{liu2024ittakestwo} introduced consistency learning across peer LiDAR representations, treating representation differences as perturbations. These approaches remain post-hoc, adjusting pseudo-label usage rather than improving their intrinsic quality.
\textsc{RePL} directly enhances pseudo-labels by correcting erroneous pseudo-labels, providing improved supervision.
\label{related}
\section{Method}
A chronic issue in semi-supervised learning is the confirmation bias towards errors of pseudo-labels, which has often been handled by post-hoc strategies such as confidence filtering or loss reweighting.
In this work, we explore a different direction:
improving the quality of unreliable pseudo-labels by refinement instead of discarding them.

To this end, we
propose \textsc{RePL}, a refinement-based semi-supervised learning framework for LiDAR semantic segmentation.
\textsc{RePL}
consists of two modules: teacher–student segmentation networks and a pseudo-label refiner. The teacher network generates pseudo-labels for unlabeled data and is updated by exponential moving average of the student parameters, while the student is trained with both labeled and pseudo-labeled data. To handle unreliable pseudo-labels, the refiner identifies uncertain voxels and corrects their pseudo-labels
through masked reconstruction. The final pseudo-labels combine reliable teacher predictions with the refined output on uncertain regions.

The remainder of this section first elaborates on three training steps of \textsc{RePL}: (1) training the student network using the standard segmentation objectives on labeled data~(\cref{method1}), (2) training the pseudo-label refiner on both labeled and unlabeled data~(\cref{method2}), and (3) semi-supervised learning of the student network with the pseudo-labels improved by the refiner~(\cref{method3}).
Then a theoretical analysis is presented to establish when the refinement is beneficial and to validate
that \textsc{RePL} operates within this regime~(\cref{theory}).

\subsection{Preliminaries}
We consider training a segmentation network on a small labeled dataset and a large unlabeled dataset.
Each point cloud is voxelized into a regular grid
$X_i \in \mathbb{R}^{C \times H \times W \times L}$,
where $C$ denotes the number of feature channels and $H \times W \times L$
is the grid size. The labels are converted into voxel-wise one-hot tensors $Y_i \in \{0,1\}^{K \times H \times W \times L}$, where $K$ is the number of classes.
The labeled and unlabeled datasets are denoted as
$D_L=\{(X_i,Y_i)\}_{i=1}^{N_l}$ and
$D_U=\{X_j\}_{j=1}^{N_u}$, respectively.

\subsection{Supervised Learning with Labeled Data}\label{method1}
Following prior work~\cite{cylindernet,kong2023lasermix,li2023less,liu2024ittakestwo}, we train the student network $f(\cdot)$ on labeled data with two complementary segmentation losses: cross-entropy for voxel-wise classification and Lovász-Softmax~\cite{berman2018lovasz} for direct IoU optimization. We denote the set of voxel grid indices as $\Omega = \{1,\dots,H\}\times\{1,\dots,W\}\times\{1,\dots,L\}$. For input $X_i$, the network outputs predictions $P_i=f(X_i)\in\mathbb{R}^{K\times H\times W\times L}$, where $[P_i]_{k,\omega}$ denotes the probability of class $k$ at voxel $\omega\in\Omega$. The ground-truth label is denoted by $[Y_i]_{k,\omega}\in\{0,1\}$.

The voxel-wise cross-entropy loss is defined as
\begin{align}
    \mathcal{L}_{\text{ce}}(P_i,Y_i)
    = -\frac{1}{|\Omega|}\sum_{\omega\in\Omega}\sum_{k=1}^{K}
    [Y_i]_{k,\omega}\,\log [P_i]_{k,\omega}.
\end{align}
For Lovász-Softmax, we first define the one-versus-rest error for each class $k$:
\begin{align}
    \mathbf{e}_i^{(k)}(\omega) = (1-[P_i]_{k,\omega}) \cdot \1_{\{[Y_i]_{k,\omega}=1\}} + [P_i]_{k,\omega} \cdot \1_{\{[Y_i]_{k,\omega} \neq 1\}}.
\end{align}
The multiclass Lovász extension is then given by
\begin{align}
    \mathcal{L}_{\text{ls}}(P_i,Y_i) = \frac{1}{|C'_i|}\sum_{k\in C'_i} \overline{\Delta}_{\text{Jacc}}\big(\mathbf{e}_i^{(k)}\big),
\end{align}
where $C'_i=\{k\mid \sum_{\omega\in\Omega}[Y_i]_{k,\omega}>0\}$ and $\overline{\Delta}_{\text{Jacc}}$ denotes the Lovász extension~\cite{berman2018lovasz} of the Jaccard loss.
The supervised learning objective combines both terms:
\begin{align}
    \mathcal{L}_{\text{ssup}} = \frac{1}{N_l}\sum_{i=1}^{N_l} \Bigl \{ \mathcal{L}_{\text{ce}}(P_i,Y_i) + \lambda_{\text{ls}}\,\mathcal{L}_{\text{ls}}(P_i,Y_i) \Bigr \}. \label{eq:sup_obj}
\end{align}

\subsection{Training Pseudo-label Refiner }\label{method2}
We employ the teacher-student framework where
the teacher $f^\tau$ network outputs predictions for unlabeled data, $Q=f^\tau(X)\in \mathbb{R}^{K \times H \times W \times L}$, as their pseudo-labels
and is updated by exponential moving average of the student network $f$~\cite{MeanTeacher}. The pseudo-label refiner network $g(\cdot)$
aims to identify unreliable teacher's predictions and refine them to improve the quality of the pseudo-labels.

\paragraph{Unreliable Voxel Identification.}\label{ed}
Voxels with unreliable pseudo-labels are identified by the agreement between the student's and teacher's predictions along with their confidence levels. For each voxel, we compute the confidence score as the maximum prediction probability across all classes for each of the two networks.
We also establish adaptive confidence thresholds using the $(100-\kappa)$-th percentile of all confidence scores within each scene, where $\kappa$ controls the strictness of the confidence requirement.
A voxel is considered reliable only if all the following conditions hold: (1) the student and teacher predict the same class, (2) the student’s confidence exceeds its adaptive threshold, and (3) the teacher’s confidence exceeds its adaptive threshold. All other voxels are treated as unreliable and marked for refinement.
The error candidate mask is then given by $
    M = \mathbf{1} - \mathbf{1}_{\text{rel}} \in \{0,1\}^{H\times W\times L},$
where $\mathbf{1}_{\text{rel}}$ denotes the
binary
mask indicating voxels that satisfy all the three conditions above.
To prevent the refiner from overfitting to error-prone regions and develop a better contextual understanding rather than simply
memorizing patterns, we also apply additional random masking $R \sim \text{Bernoulli}(\sigma)$ and define the final mask as $\tilde{M} = M \lor R$.

\paragraph{Masked Reconstruction.}
Once unreliable voxels are identified through the error candidate mask, the next step is to correct their predictions.
The core idea is to mask out these uncertain predictions and
reconstruct them through a masked reconstruction process~\cite{mae}.
Specifically, we form
masked predictions $\bar{Q}$
by substituting the unreliable predictions in $Q$ with a learnable mask token
$T\in\mathbb{R}^K$, which serves as a placeholder indicating that the original prediction at its position has been masked and needs to be reconstructed by the refiner: $\bar{Q} = (\mathbf{1}-\tilde{M}) \odot Q + \tilde{M} \odot T,$
where $\odot$ denotes element-wise multiplication.
Analogous to the mask token in masked autoencoders~\cite{mae}, $T$ is jointly optimized during training, allowing the network to learn an effective representation for masked regions. The refiner then
takes channel-wise concatenation of $X$ and $\bar{Q}$ as input and outputs refined predictions: $\hat{Q}=g(X,\bar{Q})$.

\paragraph{Training Refiner on Labeled Data.}
On labeled data, the refiner is trained to reconstruct ground-truth labels of the masked predictions. The loss for training the refiner on labeled data, $\mathcal{L}_{\text{rsup}}$, is the same as the supervised learning objective $\mathcal{L}_\text{ssup}$ in \cref{eq:sup_obj}, except that it is applied to only the masked regions (\ie, $\forall \omega \in \Omega \;\text{ s.t. }\; [\tilde{M}_i]_\omega = 1$) of the refined predictions $\hat{Q}_i$.

\paragraph{Training Refiner on Unlabeled Data.}
On unlabeled data, the refiner is further trained with a negative learning signal~\cite{negative}.
Rather than enforcing hard pseudo-labels, we suppress predictions on implausible classes by taking the teacher's top-$k$ predictions as plausible candidates and encouraging the refiner to assign low probability to the remaining classes.
This negative learning strategy offers reliable supervisory signals even when pseudo-supervision may not be sufficiently accurate on unlabeled data.
Formally, the negative learning loss is defined as the average cross-entropy penalty over unlabeled scenes in $D_U$:
\begin{align}
    \mathcal{L}_{\text{runl}} = \frac{1}{N_u}\sum_{j=1}^{N_u} \frac{1}{|\Omega|}\sum_{\omega\in\Omega} \frac{1}{|\mathcal{N}_j(\omega)|}\sum_{k\in\mathcal{N}_j(\omega)} \big\{-\log(1-[\hat{Q}_j]_{k,\omega})\big\},
\end{align}
where $\mathcal{N}_j(\omega)$ denotes the set of implausible classes for voxel $\omega$.

\paragraph{Training Refiner on Mix of Labeled and Unlabeled Data.}
Training the refiner benefits from
diverse
prediction errors of the teacher, but the limited labeled data restrict
such diversity.
To strengthen the supervision, we mix labeled and unlabeled scenes so that the refiner reconstructs labels under richer variability in distance, geometry, and density.
This produces augmented predictions with higher error rates, and allows the refiner to partially experience errors on unlabeled data.
We adopt LaserMix~\cite{kong2023lasermix} with a single inclination plane to fuse labeled and unlabeled scans at ratio $r\in(0,1)$.
Let a mix of labeled and unlabeled scenes be indexed by $m\in\{1,\dots,N_m\}$.
Given a labeled scene $(X_i, Y_i)$ and an unlabeled one $X_j$
for the $m$-th mix, LaserMix generates a selector mask $S_m \in \{0,1\}^{H \times W \times L}$ and produces the mixed input and output
$(X_m, Y_m) = (S_m \odot X_i + (\mathbf{1} - S_m) \odot X_j, S_m \odot Y_i)$. We then compute student and teacher predictions on $X_m$, $P_m = f(X_m)$ and $Q_m = f^{\tau}(X_m)$, respectively.
Following the unreliable prediction identification procedure in \cref{ed}, we construct the error mask $\tilde{M}_m$ restricted to the labeled prediction.
The supervised learning losses are then applied to voxels of the labeled scene marked as unreliable, \ie, $\forall\omega \in  \Omega \;\text{s.t.}\;  [S_m]_\omega =  [\tilde{M}_m]_\omega = 1$:
\begin{align}
    \mathcal{L}_{\text{rmix}} = \frac{1}{N_m}\sum_{m=1}^{N_m} \left\{\mathcal{L}_{\text{ce}}(\hat{Q}_m, Y_m) + \lambda_{\text{ls}} \mathcal{L}_{\text{ls}}(\hat{Q}_m, Y_m)\right\}.
\end{align}

\paragraph{Total Training Objective for Refiner.}
Each iteration optimizes the pseudo-label refiner with the summation of the three losses, $\mathcal{L}_{\text{rsup}} + \mathcal{L}_{\text{runl}} + \mathcal{L}_{\text{rmix}}$, without balancing hyper-parameters.

\subsection{Semi-supervised Learning with Pseudo-label Refiner}\label{method3}
The student network leverages the refined pseudo-labels from the refiner to improve the quality of supervision when learning with unlabeled data.

\paragraph{Pseudo-label Refinement.}
For an unlabeled input $X_j \in D_U$, we obtain predictions from both student and teacher $(P_j, Q_j)$ and construct the error mask $M_j$ as before but without random masking.
The refined pseudo-label $\tilde{Y}_j$ is then generated voxel-wise: reliable voxels follow the teacher’s prediction, while unreliable ones are replaced with the refiner’s output.
Specifically, we form masked predictions
$\bar{Q}_j$ by substituting the unreliable predictions in $Q_j$ with a learnable mask token $T$:
$\bar{Q}_j = (\mathbf{1}-M_j)\odot Q_j + M_j\odot T$
and obtain refined predictions $\hat{Q}_j = g(X_j,\bar{Q}_j)$, which are combined with teacher predictions on reliable voxels.
The refined pseudo-label $\tilde{Y}_j$ is then obtained by choosing the best class per voxel.

\paragraph{Semi-supervised Learning of Student.}
The student network is trained with three objectives, the supervised learning objective applied to labeled data (\ie, $\mathcal{L}_{\text{ssup}}$ in \cref{method1}), and two additional objectives for
learning with unlabeled data.
For the segmentation losses on unlabeled data, as the refined pseudo-labels are used directly without additional filtering, we adopt the symmetric cross-entropy~\cite{sce} instead of the standard cross-entropy to ensure more stable training. The symmetric formulation is more robust to potential noise in pseudo-labels as it penalizes over-confident predictions and provides regularization through bidirectional loss computation. Together with the Lovász-Softmax loss, this objective is applied to all voxels using the refined pseudo-labels:
\begin{align}
    \mathcal{L}_{\text{sunl}} = \frac{1}{N_u}\sum_{j=1}^{N_u} \left\{\tfrac{1}{2}\left(\mathcal{L}_{\text{ce}}(P_j,\tilde{Y}_j)+\mathcal{L}_{\text{ce}}(\tilde{Y}_j,P_j)\right) + \lambda_{\text{ls}}\, \mathcal{L}_{\text{ls}}(P_j,\tilde{Y}_j)\right\}.
\end{align}
In addition to training with pseudo-labels, we augment training by mixing a labeled scene $(X_i,Y_i)$ with an unlabeled scene $X_j$ paired with pseudo-labels $\tilde{Y}_j$. We employ the LaserMix operator with a single inclination plane. This strategy combines clean supervision from labeled data with broader coverage from pseudo-labels, exposing the student to reliable signals and diverse structures.
Given the selector mask $S_m$ by LaserMix, the mixed input and target are defined by combining ground-truth labels in the labeled region and refined pseudo-labels in the unlabeled region:
\begin{align}
    (X_m, \tilde{Y}_m) = \left(S_m \odot X_i + (\mathbf{1}-S_m)\odot X_j,\; S_m \odot Y_i + (\mathbf{1}-S_m)\odot \tilde{Y}_j \right).
\end{align}
For the mixed input $X_m$, the student network produces predictions $P_m = f(X_m)$. The loss for the mixed sample is then computed by
\begin{align}
    \mathcal{L}_{\text{smix}} = \frac{1}{N_m}\sum_{m=1}^{N_m} \left\{\tfrac{1}  {2}\left(\mathcal{L}_{\text{ce}}(P_m,\tilde{Y}_m) +\mathcal{L}_{\text{ce}}(\tilde{Y}_m, P_m)\right) +\lambda_{\text{ls}}\mathcal{L}_{\text{ls}}(P_m,\tilde{Y}_m)\right\}.
\end{align}

\paragraph{Total Training Objective for Student.}
In each iteration, the student network is optimized with the summation of the three losses, $\mathcal{L}_{\text{ssup}} + \mathcal{L}_{\text{sunl}} + \mathcal{L}_{\text{smix}}$, with no balancing hyper-parameter. The student network is optimized jointly with the pseudo-label refiner. We stop gradients between their optimization paths to prevent interference.

\subsection{Theoretical Analysis}
\label{theory}
This section rigorously analyzes if the pseudo-label refinement is truly helpful.
We first show whether the pseudo-label refinement is easier than generating high-quality pseudo-labels from scratch in \cref{pp1}.
\begin{proposition}[Refinement Difficulty]\label{pp1}
    Consider two segmentation tasks, the original task $Z: X \rightarrow Y$ and the refinement task $Z': (X, T) \rightarrow Y$, where $X$ denotes input 3D LiDAR point data, $Y$ denotes segmentation labels, and $T$ represents additional information such as teacher predictions $f(X)$.
    The difficulties of the two tasks $D(Z)$ and $D(Z')$ hold the following inequality:
    \begin{align}
        D(Z') \;=\; H(Y \mid X, T) \;\leq\; H(Y \mid X) \;=\; D(Z).
    \end{align}
\end{proposition}
This result implies that the refinement may have potential for improving pseudo-label quality.
Next, we derive a practical condition required for net performance gains in \cref{pp2}.
\begin{proposition}[Improvement Condition]\label{pp2}
    For the $j$-th scene, an error-candidate mask $M_j$ divides the voxel grid into an unreliable region $E_j$ and a reliable region $C_j$.
    Let $\pi_j$ denote the precision of this mask, which is the fraction of voxels misclassified by the teacher in $E_j$.
    The refiner operates on $E_j$, where $q_j$ and $r_j$ denote the rates at which it corrects misclassifications and incorrectly changes correct predictions, respectively.
    Then, if $q_j + r_j > 0$, the refinement improves the accuracy for scene $j$ if and only if
    \begin{align}
        \zeta_j := \pi_j - \frac{r_j}{q_j + r_j} \;>\; 0.
        \label{eq:threshold1}
    \end{align}
\end{proposition}
This proposition characterizes the trade-off between error correction and error introduction,
and its conclusion is a condition that is mild and easily satisfied by \textsc{RePL} even with the simple error estimation method in \cref{method3}, as will be demonstrated below.
The proofs for the two propositions are presented in the supplementary material.

\begin{figure}[t]
    \centering
    \begin{subfigure}{0.48\linewidth}
        \centering
        \includegraphics[width=\linewidth]{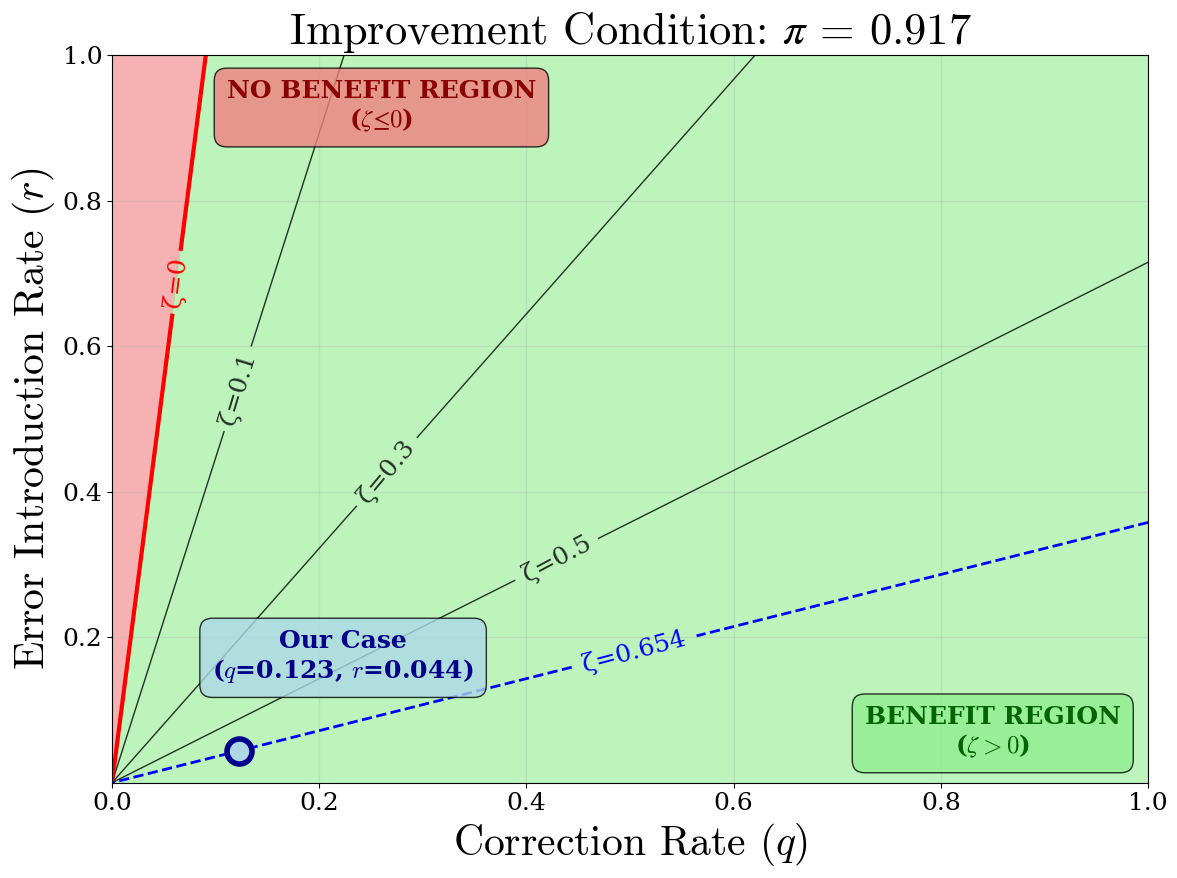}
    \end{subfigure}\hfill
    \begin{subfigure}{0.48\linewidth}
        \centering
        \includegraphics[width=\linewidth]{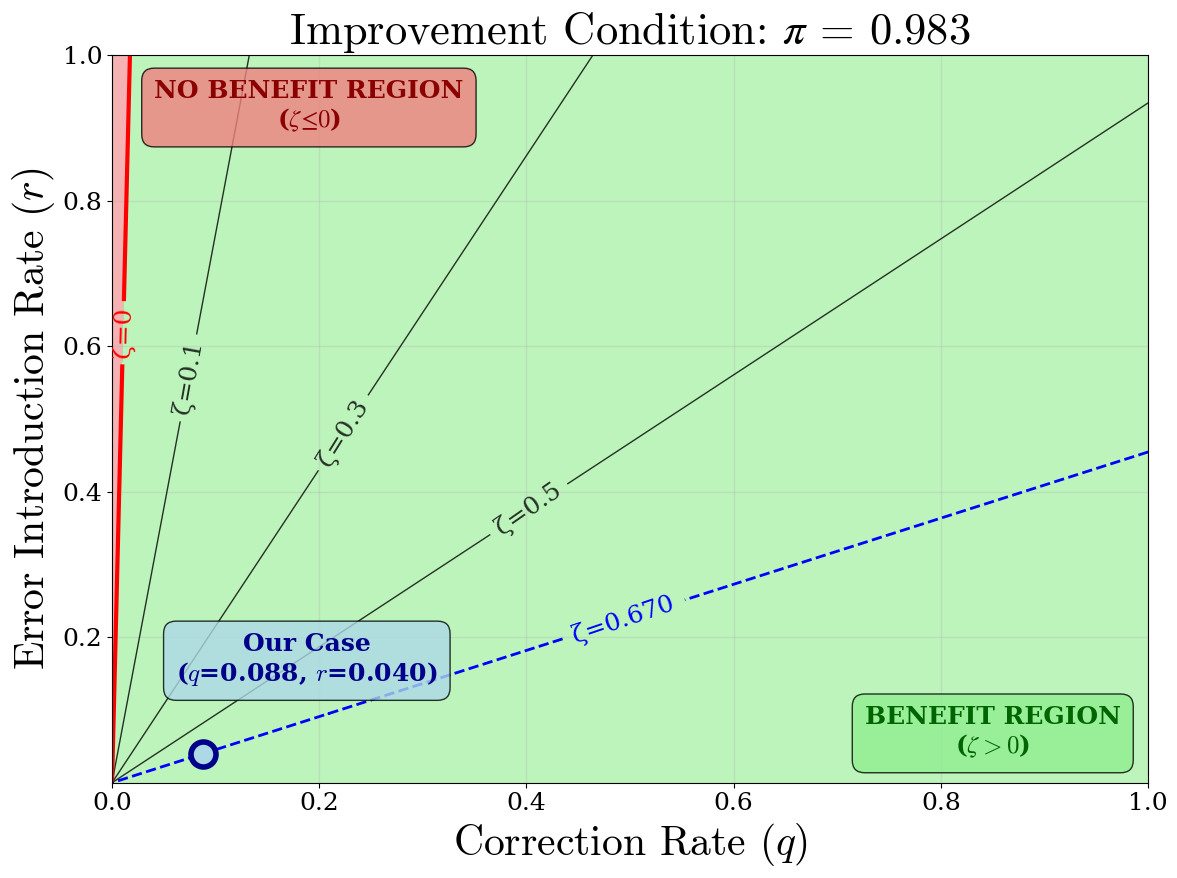}
    \end{subfigure}
    \caption{Visualization of the improvement condition from \cref{eq:threshold1} on the SemanticKITTI dataset given $\pi$. The two values of $\pi$ were derived during the actual experiments on the dataset: 1\% labeled data ($\pi = 0.917$) and 50\% labeled data ($\pi = 0.983$). Green areas indicate combinations of $q$ and $r$ where the refinement yields a net benefit ($\zeta > 0$), while red areas show such combinations leading to detriment ($\zeta \leq 0$). The results suggest that the refinement remains beneficial across a broad range of $q$ and $r$, and that \textsc{RePL} clearly helps improve the quality of pseudo-labels.}
    \label{fig:improvement_condition}
    \vspace{-3ex}
\end{figure}

\paragraph{Empirical Analysis on the Improvement Condition.}
To examine the practical implication of the condition in \cref{eq:threshold1}, we analyze the relationship between the averaged correction rate $q$ and the averaged error introduction rate $r$ using experimental results on the SemanticKITTI dataset. We consider two scenarios:
labeling 1\% of training data yielding $\pi = 0.917$, and labeling 50\% achieving $\pi = 0.983$.
\cref{fig:improvement_condition} visualizes combinations of $q$ and $r$ where the refinement yields a net benefit ($\zeta > 0$) versus those where it does not ($\zeta \leq 0$).
The results suggest that the refinement remains beneficial across a broad range of $q$ and $r$.
For instance, in the case of $\pi = 0.917$, the refinement remains beneficial as long as the error introduction rate stays below approximately \textit{eleven times} the correction rate ($r < 11.05 \cdot q$), allowing the refinement to be effective across a broad range of performance levels.
In both cases, \textsc{RePL} falls within the benefit regions ($\zeta > 0$), showing that it satisfies the theoretical condition for the improvement of pseudo-label quality
even with the simple error estimation strategy based on teacher-student agreement.
\label{method}
\section{Experiments}

\subsection{Experimental Setup}

\noindent\textbf{Datasets.}
We trained and evaluated our model on two outdoor LiDAR semantic segmentation benchmarks: nuScenes-lidarseg~\cite{nuscenes-lidarseg} and SemanticKITTI~\cite{semantickitti}. nuScenes-lidarseg is a large-scale outdoor LiDAR dataset with point-wise annotations for 16 classes. We adopted the official split of 700 sequences for training and 150 for validation, which resulted in 28,130 training and 6,019 validation point cloud scenes. SemanticKITTI is a LiDAR segmentation benchmark with 19 classes. It consists of 22 sequences, of which 10 were used for training, and 1 for validation, yielding 19,130 training and 4,071 validation scenes.

\noindent\textbf{Network Architecture.}
Following previous work~\cite{kong2023lasermix, liu2024ittakestwo, liu2025exploring}, we used Cylinder3D~\cite{cylindernet} for both the segmentation models and pseudo-label refiner, fixing the intermediate layer at 16 dimensions as specified by \cite{kong2023lasermix} and \cite{liu2024ittakestwo}.

\noindent\textbf{Implementation Details.} Our method was implemented in PyTorch~\cite{pytorch}, and trained on 8 NVIDIA RTX 6000 Ada GPUs with AdamW~\cite{adamw} and weight decay of 1e-3.
The batch size was 8 on nuScenes-lidarseg and 4 on SemanticKITTI. The learning rate was 5e-3 with cosine annealing for both the segmentation network and pseudo-label refiner. The weight update ratio $\alpha$ was 0.994. Following \cite{liu2024ittakestwo}, the loss coefficient $\lambda_{\text{ls}}$ was set to 3.0. We set the confidence percentile $\kappa$ to 40\%, $k$ of top-$k$ classes for negative learning to 3, random masking probability $\sigma$ to 0.15, and mixed data at a mixing ratio $r$ of 0.7 in the selector mask $S$.
Note that all hyper-parameters except the batch size were set identically across the two benchmarks.
\vspace{-1ex}
\begin{table}[t]
    \centering
    \caption{Comparison of different semi-supervised learning methods on nuScenes-lidarseg and SemanticKITTI while varying the ratio of labeled data.
        The best results in each column are shown in bold, and the second best are \uline{underlined}.
        Backbones marked with an asterisk (*) indicate that additional representation learning or knowledge distillation from external sources has been applied.}
    \label{mainresult1}
    \vspace{-2ex}

    \setlength{\tabcolsep}{8pt}       %

    \resizebox{\linewidth}{!}{%
        \begin{tabular}{l|c|ccccc|ccccc|}
            \toprule
            \multirow{2}{*}{Method}                          & Backbone              & \multicolumn{5}{c|}{nuScenes-lidarseg\tiny{~\cite{nuscenes-lidarseg}}} & \multicolumn{5}{c|}{SemanticKITTI\tiny{~\cite{semantickitti}}}                                                                                                                                        \\
                                                             &                       & 1\%                                                                    & 10\%                                                           & 20\%           & 50\%           & Avg.           & 1\%           & 10\%           & 20\%           & 50\%           & Avg.           \\ \midrule
            \it{Sup-only}                                    & Cylinder3D            & 50.9                                                                   & 65.9                                                           & 66.6           & 71.2           & 63.7           & 45.4          & 56.1           & 57.8           & 58.7           & 54.5           \\ \midrule
            Seal\tiny{~\cite{liu2023segment}}                & MinkUNet$^\text{*}$   & 45.8                                                                   & 63.0                                                           & -              & -              & -              & 46.6          & -              & -              & -              & -              \\
            SuperFlow\tiny{~\cite{xu20244d}}                 & MinkUNet$^\text{*}$   & 48.1                                                                   & 64.5                                                           & -              & -              & -              & 48.4          & -              & -              & -              & -              \\
            SLidR\tiny{~\cite{sautier2022image}}             & Cylinder3D$^\text{*}$ & 39.0                                                                   & 58.8                                                           & -              & -              & -              & 44.6          & -              & -              & -              & -              \\ \midrule
            Lim3D\tiny{~\cite{li2023less}}                   & LiM3D                 & -                                                                      & -                                                              & -              & -              & -              & 58.4          & 59.5           & 63.1           & 63.6           & 61.2           \\ \midrule
            MT\tiny{~\cite{MeanTeacher}}                     & Cylinder3D            & 51.6                                                                   & 66.0                                                           & 67.1           & 71.7           & 64.1           & 45.4          & 57.1           & 59.2           & 60.0           & 55.4           \\
            CBST\tiny{~\cite{cbst}}                          & Cylinder3D            & 53.0                                                                   & 66.5                                                           & 69.6           & 71.6           & 65.2           & 48.8          & 58.3           & 59.4           & 59.7           & 56.6           \\
            CPS\tiny{~\cite{cps}}                            & Cylinder3D            & 52.9                                                                   & 66.3                                                           & 70.0           & 72.5           & 65.4           & 46.7          & 58.7           & 59.6           & 60.5           & 56.4           \\
            LaserMix\tiny{~\cite{kong2023lasermix}}          & Cylinder3D            & 55.3                                                                   & 69.9                                                           & 71.8           & 73.2           & 67.6           & 50.6          & 60.0           & 61.9           & 62.3           & 58.7           \\
            IT2\tiny{~\cite{liu2024ittakestwo}}              & Cylinder3D            & {\uline{57.5}}                                                         & {\uline{72.1}}                                                 & {\uline{73.5}} & {\uline{74.1}} & {\uline{69.3}} & 52.0          & 61.4           & 62.1           & 62.5           & 59.5           \\
            AIScene\tiny{~\cite{liu2025exploring}}           & Cylinder3D            & 56.6                                                                   & 70.2                                                           & 72.8           & 73.9           & 68.4           & 54.5          & \textbf{63.3}  & \textbf{63.7}  & {\uline{64.3}} & {\uline{61.5}} \\
            FrustrumMix\tiny{~\cite{xu2025frnet}}            & Cylinder3D            & \textbf{60.0}                                                          & 70.0                                                           & 72.6           & \uline{74.1}   & 69.2           & \uline{55.7}  & {\uline{62.5}} & 63.0           & 64.9           & {\uline{61.5}} \\
            LaserMix++\tiny{~\cite{kong2025multi}}           & Cylinder3D            & 58.5                                                                   & 71.1                                                           & 72.8           & 74.0           & 69.1           & \textbf{56.2} & 62.3           & 62.9           & 63.4           & 61.2           \\ \midrule
            \rowcolor{gray!20} \textbf{\textsc{RePL}~(Ours)} & Cylinder3D            & \textbf{60.0}                                                          & \textbf{74.4}                                                  & \textbf{75.0}  & \textbf{75.8}  & \textbf{71.3}  & 54.7          & {\uline{62.5}} & {\uline{63.2}} & \textbf{65.9}  & \textbf{61.6}  \\ \bottomrule
        \end{tabular}
    }
    \vspace{-4ex}
\end{table}

\subsection{Comparison with State of the Art}
\begin{figure}[t]
    \centering
    \includegraphics[width=0.9\linewidth, height=0.55\linewidth, trim=0 5 0 0]{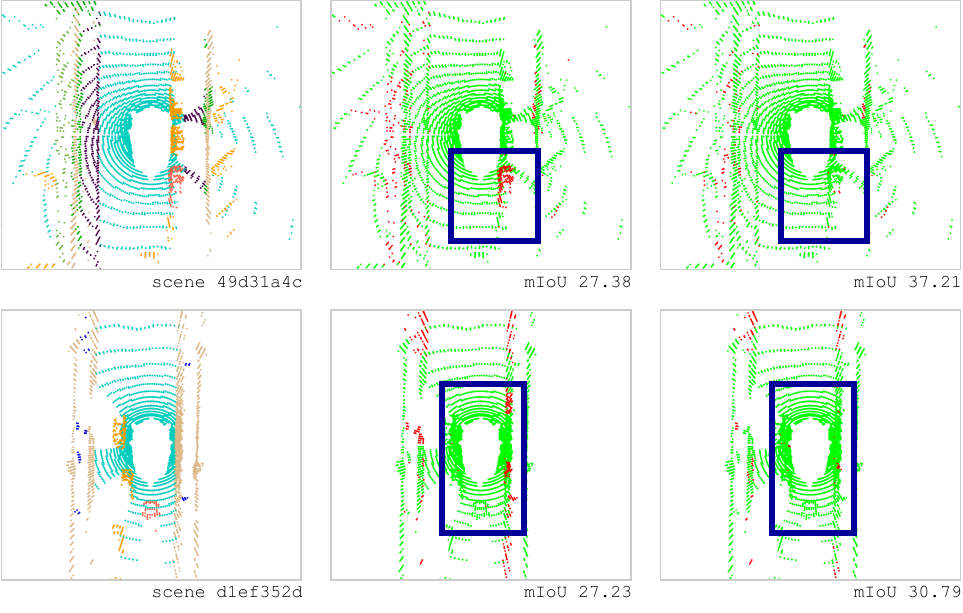}
    \begin{subfigure}{0.285\linewidth}
        \centering
        \caption{Ground-truth}
    \end{subfigure}
    \begin{subfigure}{0.33\linewidth}
        \centering
        \caption{Initial Pseudo-labels}
    \end{subfigure}
    \begin{subfigure}{0.285\linewidth}
        \centering
        \caption{Refined Pseudo-labels}
    \end{subfigure}
    \caption{
        Qualitative results of refined pseudo-labels and their initial predictions on the unlabeled set of nuScenes-lidarseg, at the end of training. Correct and incorrect predictions are shown in green and red, respectively. The model was trained with a 1\% label ratio.
    }
    \label{fig:qual}
    \vspace{-3ex}
\end{figure}

\textsc{RePL} was compared with latest methods using Cylinder3D as the backbone, namely AIScene~\cite{liu2025exploring}, IT2~\cite{liu2024ittakestwo}, FrustrumMix~\cite{xu2025frnet}, and conventional methods employing semi-supervision~\cite{MeanTeacher, cbst, cps, kong2023lasermix, kong2025multi}.
For a more comprehensive comparison, we further evaluated against Seal~\cite{liu2023segment}, SuperFlow~\cite{xu20244d}, and SLidR~\cite{sautier2022image}, which leverage external sources or additional representation learning, and Lim3D~\cite{li2023less}, which uses a distinct backbone based on Cylinder3D.
\Cref{mainresult1} summarizes the results on the validation sets of nuScenes-lidarseg~\cite{nuscenes-lidarseg} and SemanticKITTI~\cite{semantickitti} with various ratio of labeled data: 1\%, 10\%, 20\%, and 50\%. \textit{Sup-only} denotes the baseline performance of training only with labeled data.
On nuScenes-lidarseg, \textsc{RePL} outperformed all competing methods.
Compared to IT2, the second-best, it achieved an average gain of +2.0 in mIoU.
\textsc{RePL} also showed strong results on SemanticKITTI, achieving the best in average mIoU. These results are particularly impressive regarding that \textsc{RePL} fixes hyperparameters across both benchmarks, unlike the strong competitors like AIScene~\cite{liu2025exploring} and FrustrumMix~\cite{xu2025frnet} that tune key hyperparameters per benchmark.
\Cref{scribble_result} shows the results on the ScribbleKITTI~\cite{weak1} benchmark, which uses scribble-based weak annotations as labeled supervision. \textsc{RePL} achieved the highest average mIoU and outperformed all competing methods at 1\%, 10\%, and 50\% labeled data ratios.
\Cref{fig:qual} qualitatively compares pseudo-labels before and after the refinement by \textsc{RePL} at the end of training on the unlabeled data of nuScenes-lidarseg.

\subsection{In-depth Analysis}

This section studies the contribution of each component of \textsc{RePL}, and investigates the aspects of the pseudo-label refinement in details.
All experiments were conducted on the validation set, except for the pseudo-label refinement analysis, which used the unlabeled training data.

\noindent\textbf{Ablation Study on Loss Components.} We assessed the contribution of individual loss terms by incrementally adding them for the refiner and segmentation network. For the refiner~(\Cref{tab:ablation_refiner}), the supervised-only baseline yielded 50.9 mIoU. Adding $\mathcal{L}_\text{rsup}$ improved performance to 57.2 mIoU, and including $\mathcal{L}_\text{runl}$ further raised it to 58.7 mIoU. With all three objectives, including $\mathcal{L}_\text{rmix}$, accuracy reached 60.0 mIoU, confirming their complementary effect. The averaged improvement condition $\zeta$ also consistently increased as each loss component is added.
For the segmentation network~(\Cref{tab:ablation_segmentation}), the supervised-only baseline also scored 50.9 mIoU. Adding the semi-supervised loss, $\mathcal{L}_\text{sunl}$ , improved performance to 58.1 mIoU, while its variant without symmetric cross-entropy (\ding{115}) with $\mathcal{L}_{\text{smix}}$ gave 58.0 mIoU. Using all three training objectives yielded the best result of 60.0 mIoU, highlighting the benefit of jointly optimizing supervised learning on labeled data, semi-supervised learning with refined pseudo-labels, and mixed scene training.

\begin{table*}[t]
    \centering
    \begin{minipage}[t]{0.48\linewidth} %
        \centering
        \caption{
            Comparison of different semi-supervised learning methods on ScribbleKITTI while varying the ratio of labeled data. The best results are in bold, and the second best \uline{underlined}.
        }
        \label{scribble_result}
        \vspace{-2ex}
        \resizebox{\linewidth}{!}{%
            \renewcommand{\arraystretch}{0.88}
            \setlength{\tabcolsep}{5pt}
            \begin{tabular}{l|ccccc}
                \toprule
                Method                                           & 1\%           & 10\%          & 20\%          & 50\%          & Avg.          \\ \midrule
                \it{Sup-only}                                    & 39.2          & 48.0          & 52.1          & 53.8          & 48.3          \\ \midrule
                MT\tiny{~\cite{MeanTeacher}}                     & 41.0          & 50.1          & 52.8          & 53.9          & 49.5          \\
                CBST\tiny{~\cite{cbst}}                          & 41.5          & 50.6          & 53.3          & 54.5          & 50.0          \\
                CPS\tiny{~\cite{cps}}                            & 41.4          & 51.8          & 53.9          & 54.8          & 50.5          \\
                LaserMix\tiny{~\cite{kong2023lasermix}}          & 44.2          & 53.7          & 55.1          & 56.8          & 52.5          \\
                IT2\tiny{~\cite{liu2024ittakestwo}}              & \uline{47.9}  & \uline{56.7}  & 57.5          & 58.3          & 55.1          \\
                FrustrumMix\tiny{~\cite{xu2025frnet}}            & 45.6          & 55.7          & \textbf{58.2} & \uline{60.8}  & 55.1          \\
                LaserMix++\tiny{~\cite{kong2025multi}}           & 47.3          & \uline{56.7}  & \uline{57.6}  & 59.8          & \uline{55.4}  \\ \midrule
                \rowcolor{gray!20} \textbf{\textsc{RePL}~(Ours)} & \textbf{48.7} & \textbf{57.3} & \uline{57.6}  & \textbf{61.0} & \textbf{56.1} \\ \bottomrule
            \end{tabular}
        }
    \end{minipage}
    \hfill %
    \begin{minipage}[t]{0.48\linewidth} %
        \centering
        \caption{Impact of the losses for the pseudo-label refiner.
            Each row shows the average improvement condition $\zeta$ and mean IoU when training with different subsets of the losses.}
        \label{tab:ablation_refiner}
        \setlength{\tabcolsep}{4pt}       %
        \begin{tabular}{ccc|c|c}
            \toprule
            $\mathcal{L}_{\text{rsup}}$ & $\mathcal{L}_{\text{runl}}$ & $\mathcal{L}_{\text{rmix}}$ & $\zeta$ & mIoU \\ \midrule
                                        &                             &                             & -       & 50.9 \\ \midrule
            \cmark                      &                             &                             & 0.327   & 57.2 \\
            \cmark                      & \cmark                      &                             & 0.353   & 58.7 \\
            \cmark                      & \cmark                      & \cmark                      & 0.430   & 60.0 \\ \bottomrule
        \end{tabular}
    \end{minipage}
    \renewcommand{\arraystretch}{1.0} %
    \vspace{-1ex}
\end{table*}

\begin{table*}[t]
    \centering
    \begin{minipage}[t]{0.48\linewidth}
        \centering
        \caption{Impact of the losses for learning the LiDAR semantic segmentation network.
        In $\mathcal{L}_{\text{sunl}}$, \ding{115} denotes the omission of the symmetric cross-entropy.}
        \label{tab:ablation_segmentation}
        \vspace{-2ex}
        \renewcommand{\arraystretch}{0.75} %
        \setlength{\tabcolsep}{6pt}       %
        \begin{tabular}{ccc|c}
            \toprule
            $\mathcal{L}_{\text{ssup}}$ & $\mathcal{L}_{\text{sunl}}$ & $\mathcal{L}_{\text{smix}}$ & mIoU \\ \midrule
            \cmark                      &                             &                             & 50.9 \\ \midrule
            \cmark                      & \cmark                      &                             & 58.1 \\
            \cmark                      & \ding{115}                  & \cmark                      & 58.0 \\
            \cmark                      & \cmark                      & \cmark                      & 60.0 \\ \bottomrule
        \end{tabular}
    \end{minipage}
    \hfill
    \begin{minipage}[t]{0.48\linewidth}
        \centering
        \caption{Sensitivity to the quality of the error candidate mask in LiDAR semantic segmentation accuracy. Different error mask generation strategies were compared at inference time.}
        \label{tab:comparison_sampling}
        \renewcommand{\arraystretch}{1.2}
        \vspace{-2ex}
        \resizebox{0.95\linewidth}{!}{%
            \begin{tabular}{l|ccccc|c}
                \toprule
                \multirow{2}{*}{Setting} & \multirow{2}{*}{Baseline} & \multicolumn{3}{c}{Random} & \multirow{2}{*}{Oracle} & \multirow{2}{*}{Ours}               \\ \cline{3-5}
                                         &                           & 25\%                       & 50\%                    & 75\%                  &      &      \\ \midrule
                mIoU                     & 57.0                      & 57.6                       & 58.2                    & 58.7                  & 67.3 & 60.0 \\ \bottomrule
            \end{tabular}
        }
        \renewcommand{\arraystretch}{0.9}
    \end{minipage}
    \renewcommand{\arraystretch}{1.0} %
    \vspace{-5ex}
\end{table*}

\begin{table*}[t]
    \centering
    \begin{minipage}[t]{0.48\linewidth}
        \centering
        \caption{Impact of the random masking strategy for training the pseudo-label refiner in the segmentation quality of the final model.}
        \label{tab:ablation_mask}
        \vspace{-2ex}
        \renewcommand{\arraystretch}{0.9} %
        \setlength{\tabcolsep}{10pt}       %
        \begin{tabular}{l|c}
            \toprule
            Setting                          & mIoU \\ \midrule
            w/o Random Masking               & 57.7 \\
            w/ \hspace{0.8mm} Random Masking & 60.0 \\
            \bottomrule
        \end{tabular}
    \end{minipage}
    \hfill
    \begin{minipage}[t]{0.48\linewidth}
        \centering
        \caption{Computational cost analysis on nuScenes-lidarseg.}
        \label{tab:computational_cost}
        \renewcommand{\arraystretch}{1.2}
        \resizebox{0.98\linewidth}{!}{%
            \begin{tabular}{lccc}
                \toprule
                Method             & Latency (s) & Memory (MB) & mIoU \\
                \midrule
                Baseline           & 0.43        & 1231        & 50.9 \\
                Baseline + Refiner & 0.68        & 1627        & 60.0 \\
                \midrule
                $\Delta$           & +0.25       & +396        & +9.1 \\
                \bottomrule
            \end{tabular}
        }
    \end{minipage}
    \renewcommand{\arraystretch}{1.0} %
    \vspace{-5ex}
\end{table*}

\begin{figure}[t]
    \centering
    \includegraphics[width=0.9\linewidth, height=0.55\linewidth, trim=0 5 0 0]{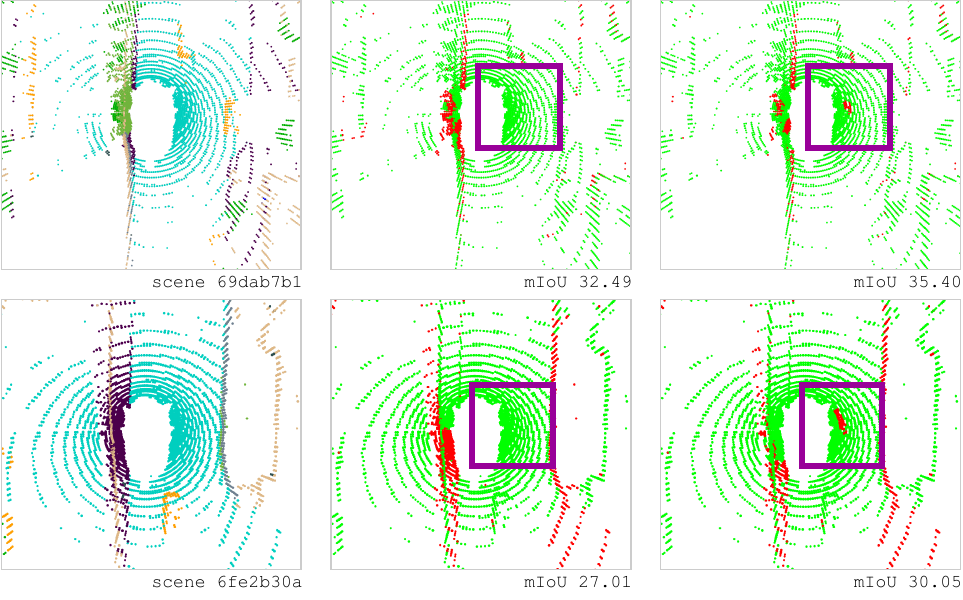}
    \begin{subfigure}{0.285\linewidth}
        \centering
        \caption{Ground-truth}
    \end{subfigure}
    \begin{subfigure}{0.33\linewidth}
        \centering
        \caption{Initial Pseudo-labels}
    \end{subfigure}
    \begin{subfigure}{0.285\linewidth}
        \centering
        \caption{Refined Pseudo-labels}
    \end{subfigure}
    \caption{Failure cases on the unlabeled set of nuScenes-lidarseg at the end of training with a 1\% label ratio. Correct and incorrect predictions are shown in green and red, respectively.}
    \label{fig:qual_fail}
    \vspace{-5ex}
\end{figure}

\begin{wrapfigure}{r}{0.48\linewidth}
    \centering
    \vspace{-5ex}
    \includegraphics[width=\linewidth]{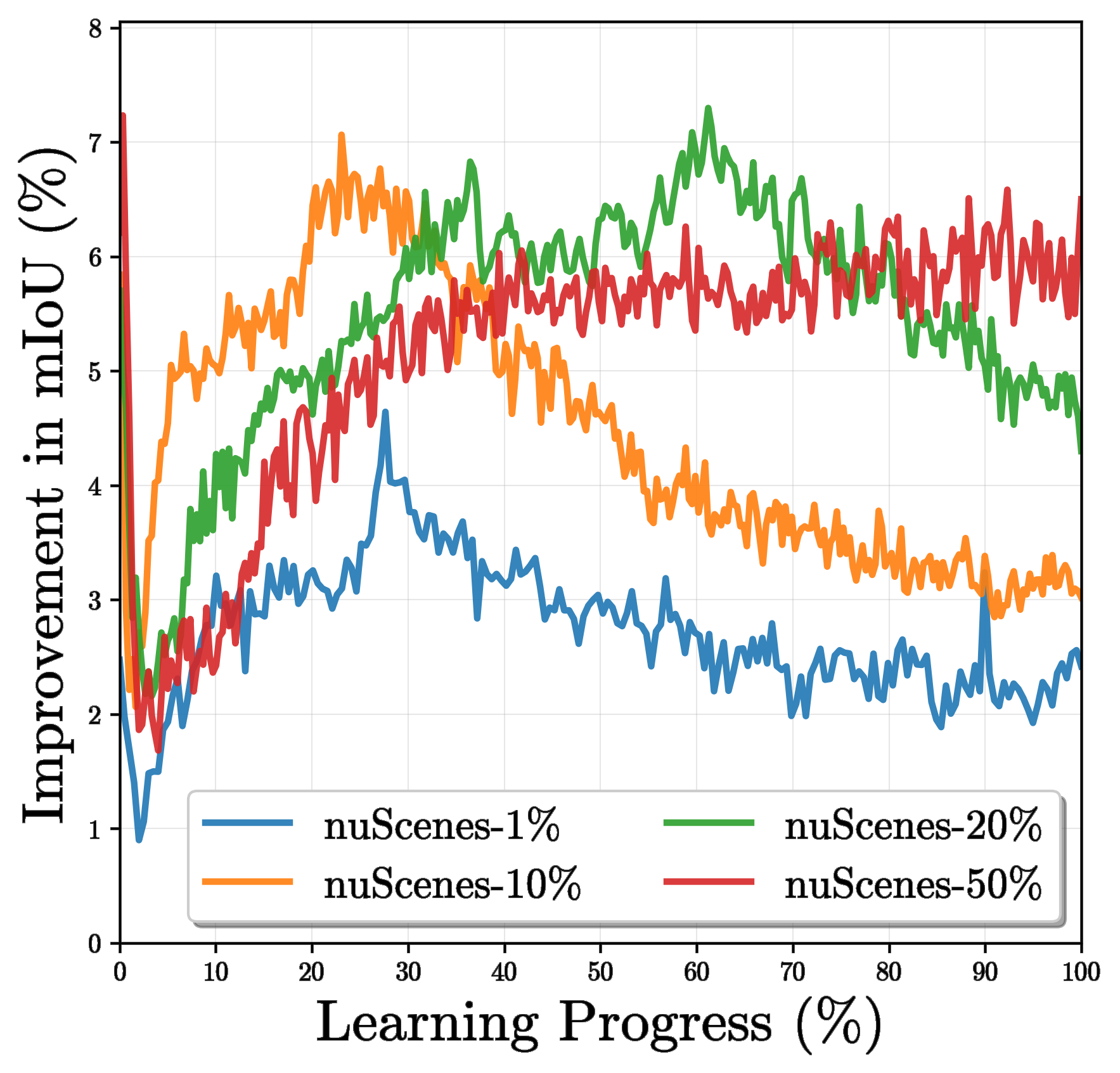}
    \caption{Pseudo-label quality improvement by the refiner during training on nuScenes-lidarseg.}
    \label{fig:nusc_improve}
    \vspace{-5ex}
\end{wrapfigure}
\noindent\textbf{Sensitivity to the Quality of Error Candidate Mask.}
We analyzed how the quality of the error candidate mask influences inference performance by replacing our heuristic error mask with different alternatives on the validation set. As shown in \Cref{tab:comparison_sampling}, random masks yielded modest improvements over the baseline (no refinement) of the teacher. Our heuristic error mask provided a clear gain, while an oracle error mask derived from ground-truth labels further improved performance to 67.3 mIoU. These results indicate that even a simple heuristic achieves competitive improvements, with more accurate error mask offering substantial room for further gains.

\noindent\textbf{Impact of the Random Masking Strategy.} We investigated whether training with random masking improves performance on the validation set. As shown in \Cref{tab:ablation_mask}, incorporating random masking yielded higher performance~(60.0 mIoU) compared to training without it (57.7 mIoU). This indicates that random masking serves as a regularizer, helping the network handle erroneous predictions more effectively and improving performance during inference.

\noindent\textbf{Analysis on Pseudo-label Quality Improvement throughout Training.}
We report the trend of pseudo-label quality improvement throughout training for different labeled data ratios (1\%, 10\%, 20\%, 50\%) on the unlabeled data of nuScenes-lidarseg in \cref{fig:nusc_improve}. During early stage of training, the improvement was relatively low across all ratios as the refiner learns error correction from scratch. As training progresses, the improvement increased as the refiner learns to correct errors more effectively. However, the improvement gradually declined in later stages as the segmentation network itself becomes accurate, leaving less room for the refiner to provide meaningful corrections to the already high-quality predictions. \textsc{RePL} showed effectiveness across all labeled data ratios, with better performance and scalability at higher ratios.

\noindent\textbf{Analysis on Computational Cost.}
To quantify the additional overhead by the refiner, we measured the latency and memory usage during inference on the validation set using a single batch. As shown in \Cref{tab:computational_cost}, the refiner adds approximately 0.25 seconds of latency and 396 MB of memory, while providing a substantial improvement of +9.1 mIoU from the supervised-only baseline. These results demonstrate that the added computational cost is moderate relative to the significant accuracy gains.

\begin{wrapfigure}{r}{0.48\linewidth}
    \centering
    \vspace{-4ex}
    \includegraphics[width=\linewidth]{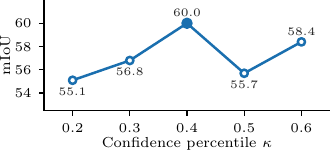}
    \caption{Sensitivity analysis of a hyper-parameter $\kappa$.}
    \label{tab:ablation_hyperparams}
    \vspace{-5ex}
\end{wrapfigure}
\noindent\textbf{Ablation Study on Unreliable Voxel Identification.}
We analyzed the sensitivity of an unreliable voxel identification in \textsc{RePL} on the validation set.
As shown in \Cref{tab:ablation_hyperparams}, the confidence percentile $\kappa = 0.4$ yields the best performance at 60.0 mIoU, while $\kappa=$ 0.2, 0.3, 0.5, and 0.6 result in suboptimal performance at 55.1, 56.8, 55.7, and 58.4 mIoU, respectively.

\noindent\textbf{Analysis on Failure Cases.}
Despite overall improvements, \textsc{RePL} occasionally introduces errors by over-correcting initially accurate predictions. \Cref{fig:qual_fail} shows representative failure cases (purple boxes). Nevertheless, the mIoU gain indicates that successful corrections outweigh these localized failures, leading to overall enhancement of the pseudo-labels.

\noindent\textbf{Discard Baseline.}
To verify that performance gains are derived from the refinement module and not from discarding uncertain voxels, we compared against a discard-only baseline that uses the same error masks as \textsc{RePL} but removes detected unreliable voxels from the pseudo-label loss instead of reconstructing them.
On nuScenes-lidarseg 1\%, this baseline achieves 53.5 mIoU, below self-training with original pseudo-labels (54.3 mIoU) and far below \textsc{RePL} (60.0 mIoU).
This indicates that the improvements are driven by the proposed refinement module.

\begin{table}[t]
    \centering
    \caption{Per-class pseudo-label quality improvement on the unlabeled set of nuScenes-lidarseg with 1\% labeled ratio.
        The column marked as \% indicates class frequency.}
    \label{tab:perclass}
    \vspace{-2ex}
    {\setlength{\tabcolsep}{4pt}
        \renewcommand{\arraystretch}{0.95}
        \resizebox{\linewidth}{!}{%
            \begin{tabular}{lrr|lrr|lrr|lrr}
                \toprule
                Class    & \%   & $\Delta$IoU & Class   & \%   & $\Delta$IoU & Class          & \%   & $\Delta$IoU & Class        & \%   & $\Delta$IoU \\
                \midrule
                barrier  & 1.10 & +3.45       & bus     & 0.55 & +1.04       & const.\ veh.\  & 0.18 & +2.54       & motorcycle   & 0.05 & +2.55       \\
                bicycle  & 0.02 & +2.68       & car     & 4.42 & +1.60       & pedestrian     & 0.27 & +6.47       & traffic cone & 0.09 & +7.86       \\
                trailer  & 0.58 & +0.60       & truck   & 1.83 & +1.02       & driv.\ surf.\  & 37.6 & $-$1.47     & flat other   & 1.02 & +2.60       \\
                sidewalk & 8.31 & $-$2.40     & terrain & 8.34 & +4.14       & manmade        & 21.1 & +4.41       & vegetation   & 14.5 & +5.35       \\
                \bottomrule
            \end{tabular}
        }}
    \vspace{-4ex}
\end{table}

\noindent\textbf{Per-class Analysis.}
As shown in \Cref{tab:perclass}, \textsc{RePL} improved pseudo-label quality for 14 of 16 classes on the unlabeled set of nuScenes-lidarseg with 1\% labeled ratio.
Rare and small classes such as traffic cone (0.09\% freq., +7.86 mIoU), pedestrian (0.27\%, +6.47), vegetation (14.5\%, +5.35), and manmade (21.1\%, +4.41) benefited the most.
Overall, the mIoU improved from 59.42 to 62.07.

\noindent\textbf{Range-based Analysis.}
To show that pseudo-label refinement is not limited to near-range regions, we measured pseudo-label quality gains across distance bins on nuScenes-lidarseg 1\%.
\textsc{RePL} improved pseudo-label quality consistently across all evaluated distance bins (0--40\,m), with gains of +1.95 (0--10\,m), +1.30 (10--20\,m), +1.34 (20--30\,m), and +1.26 (30--40\,m).
The results confirm that the refinement also benefits far-range regions.

\noindent\textbf{Generalization to Other Backbones.}
To assess whether \textsc{RePL} is tied to Cylinder3D, we replaced it with SphereFormer~\cite{lai2023spherical} without architecture-aware tuning.
On nuScenes-lidarseg 1\%, \textsc{RePL} with SphereFormer improved from the supervised-only baseline of 50.7 to 56.5 mIoU.
This demonstrates that the proposed framework is not architecture-specific.
\label{experiment}
\vspace{-1ex}\section{Conclusion}\vspace{-1ex}
We presented \textsc{RePL}, a semi-supervised learning framework for LiDAR semantic segmentation that refines pseudo-labels through a two-stage mechanism of error estimation and masked reconstruction.
The framework integrates a teacher-student segmentation network with a pseudo-label refiner to identify unreliable predictions and reconstruct them into cleaner supervision signals.
We also provided theoretical analysis establishing the mathematical conditions under which refinement improves pseudo-label quality.
With this design, our method achieved state-of-the-art results on nuScenes-lidarseg and SemanticKITTI across various label ratios.
A current limitation is that the consensus-based error estimation may miss errors on which the teacher and student are jointly overconfident.
Replacing the heuristic error mask with a learned error-localization approach such as ELN~\cite{Kwon_2022_CVPR} is a promising direction for further closing the gap to the oracle.

\noindent\textbf{Acknowledgments.}
This work was supported by the Scaleup TIPS grant (RS-2023-00321784: Development of Novel Generative AI Technology to Generate Domain-Specific Synthetic Data) and the IITP grants (%
RS-2022-II220290,
RS-2022-II220926,
RS-2024-00509258, RS-2024-00469482,
RS-2024-00457882,
RS-2019-II191906%
) funded by Ministry of Science and ICT, Korea.
\label{conclusion}
\clearpage
\bibliographystyle{splncs04}
\bibliography{main}

\end{document}